\documentclass[conference]{IEEEtran}

\usepackage{array}
\newcolumntype{P}[1]{>{\centering\arraybackslash}p{#1}}
\usepackage{color}
\usepackage{multirow}

\ifCLASSINFOpdf
  \usepackage[pdftex]{graphicx}
  \graphicspath{{./images/}}
\else
\fi

\usepackage{cite}

\begin{document}

\title{Low-Power Hardware-Based Deep-Learning Diagnostics Support Case Study}

\author{\IEEEauthorblockN{Khushal Sethi, Vivek Parmar and Manan Suri\IEEEauthorrefmark{1}}
\IEEEauthorblockA{Department of Electrical Engineering, Indian Institute of Technology, Delhi, \\Email: manansuri@ee.iitd.ac.in}}

%


\maketitle

\begin{abstract}
Deep learning research has generated widespread interest leading to emergence of a large variety of technological innovations and applications. As significant proportion of deep learning research focuses on vision based applications, there exists a potential for using some of these techniques to enable low-power portable health-care diagnostic support solutions. In this paper, we propose an embedded-hardware-based implementation of microscopy diagnostic support system for PoC case study on: (a) Malaria in thick blood smears, (b) Tuberculosis in sputum samples, and (c) Intestinal parasite infection in stool samples. We use a Squeeze-Net based model to reduce the network size and computation time. We also utilize the Trained Quantization technique to further reduce memory footprint of the learned models. This enables microscopy-based detection of pathogens that classifies with laboratory expert level accuracy as a standalone embedded hardware platform. The proposed implementation is 6x more power-efficient compared to conventional CPU-based implementation and has an inference time of $\sim$ 3 ms/sample.

\end{abstract}


%
\IEEEpeerreviewmaketitle

\section{Introduction}

Artificial intelligence and deep learning research have enabled techniques leading to development of innovative solutions for a wide variety of applications  \cite{bashir2019power, sethi2021efficient, ji2020reconfigurable,ji2020compacc, sethi2020nv, sethi2020design, sethi2018low, sethi2019optimized, radway2021future}. Diagnostic support solutions in healthcare may exploit image processing and image recognition approaches. Given recent advances in dedicated deep-learning hardware accelerators, there are possibilities to realize low-power embedded solutions for point of care diagnostic support systems in remote locations with limited network connectivity, lack of infrastructure, and difficult access to human experts.
In this paper work we focus on a low-power hardware implementation of deep learning based proof-of-concept (PoC) microscopy diagnostic support system for three different diseases: Malaria, Tuberculosis and Intestinal Parasite Infection. 


Several diagnostic studies for these diseases based on computer-vision methodologies, conventional machine learning and deep learning have been proposed in literature \cite{quinn2016deep,bates2004improving,beaver1949quantitative,quinn2014automated,yokota2002scalable,grull2011accelerating}. However, most of the implementations in literature are based on conventional CPUs, high-end GPUs or FPGAs. For low-power and portable diagnostic support solutions there is a need of exploring the potential of dedicated acceleration hardware such as special purpose ASICs.

Key contributions of this work are:
\begin{itemize}
\item We propose a SqueezeNet based advanced deep learning model that is capable of recognizing characteristics of pathogens in three different disease sample images. Compression method is applied to trained model in order to optimize memory footprint and computational complexity.
\item The proposed Convolutional Neural Network (CNN) inference model is realized on a dedicated low-power hardware platform using the Myriad VPU \cite{moloney2014myriad} (Intel Movidius NCS) coupled with Raspberry Pi.
\item Performance benchmarking of the proposed CNN on i7 CPU and the dedicated hardware VPU accelerated platform is presented.
\item The proposed PoC platform can be integrated with portable microscope for rapid detection of diseases in resource-constrained environments.
\end{itemize}

The paper is organized as follows: Section \ref{dset} discusses the datasets used for the study. Section \ref{meth} describes the proposed CNN architecture, its software and dedicated hardware implementations. Section \ref{res} discusses the results and Section \ref{conc} lists the key conclusions.


\begin{figure}[!htbp]
\centering
\includegraphics[scale=0.35]{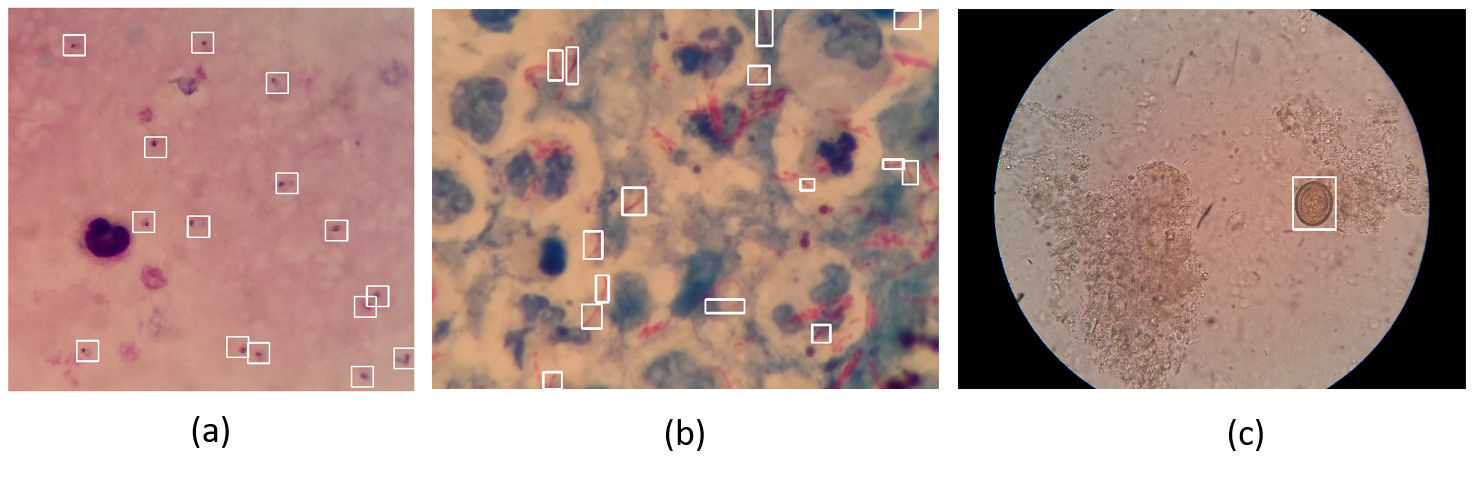}
\caption{Annotated images in (a) Malaria dataset showing plasmodium (b) Tuberculosis dataset showing red/pink bacilli (c) Intestinal Parasites dataset showing hookworm eggs\cite{quinn2016deep}. }
\label{fig:data}
\end{figure}

\section{Microscopy Based Datasets}
\label{dset}
We used the automated laboratory diagnostics dataset released by the Artificial Intelligence Research Group, Makerere University, Uganda \cite{aldair}. The malaria dataset contains images taken from thick blood smears at 1000x magnification with annotated plasmodium (7245 objects in 1182 images). The Tuberculosis dataset contains images taken from fresh sputum and stained using ZN (Ziehl Neelsen) stain at 1000x magnification with annotated tuberculosis bacilli (3734 objects in 928 images). The intestinal parasites dataset contains images taken from slides of a portion of stool sample examined under 400x magnification annotated with eggs of hookworm, Taenia and Hymenolepsis nana (162 objects in 1217 images)\cite{quinn2016deep}. Sample Annotated Images from the dataset are shown in Fig. \ref{fig:data}.

Using the images, we produced positive and negative sample images for training a binary classification model that can detect the presence of pathogen. Positive samples (i.e. those containing plasmodium, bacilli or parasite eggs respectively) were produced by taking the centered bounding boxes in the annotation. Negative samples in each image (i.e. with absence of any of these pathogens) were taken from random locations not intersecting with any annotated bounding boxes. As dominant image areas did not contain pathogen objects, the ratio of positive to negative samples was highly skewed. Thus, some negative samples were randomly discarded and new positive samples were created by applying different transformations such as rotation and flipping\cite{quinn2016deep}. The produced sample images were then down-sized to 20 x 20 (for malaria and tuberculosis) and 30 x 30 (for intestinal parasites) to reduce the computational complexity of the algorithm.

\section{Methodology and Experiments}
\label{meth}
\begin{figure}[t]
\centering
\includegraphics[scale=0.40]{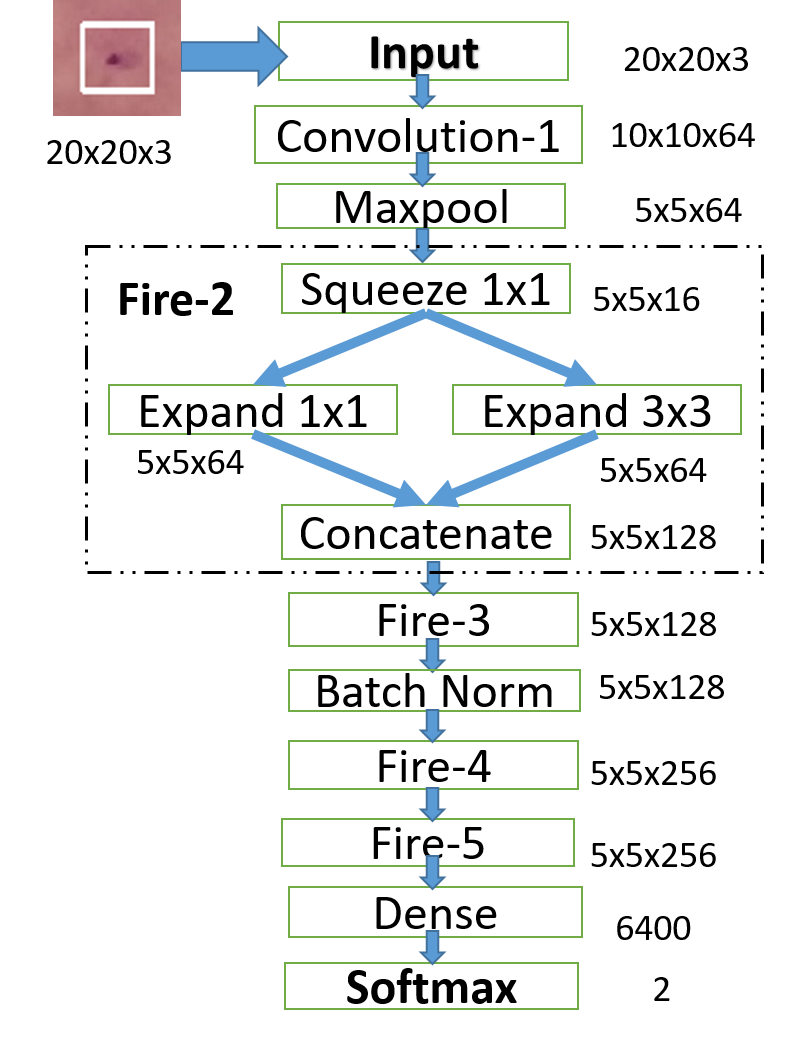}
\caption{Architecture of our Proposed Custom CNN.}
\label{fig:net}
\end{figure}
\subsection{Proposed Custom CNN Architecture}
For classification, we designed a custom convolutional neural network inspired from the SqueezeNet Architecture\cite{DBLP:journals/corr/IandolaMAHDK16}.
Architecture of the proposed network is shown in Fig \ref{fig:net}. The network layers are described as follows:
\begin{enumerate}
\item \textbf{Convolutional Layer}: These layers perform a convolution operation on the input, passing the result to the next layer. They have learnable weights and biases.  Filters of size 3 x 3 were used for the Convolution-1 layer of our network.  
\item \textbf{Max-Pooling Layer}: This layer in the network uses the maximum value in a sliding window of size 3 x 3 across the previous layer.
\item \textbf{Fire module}: This layer is comprised of - a \emph{squeeze} convolution layer \cite{DBLP:journals/corr/IandolaMAHDK16} (which uses 1 x 1 point-wise filters), feeding into an \emph{expand} layer that has a mix of 1 x 1 and 3 x 3 convolution filters.
\item \textbf{Batch Normalization}: This layer is used to reduce the impact of previous layers on learning after each batch by applying a transformation that maintains the mean activation close to 0 and the activation standard deviation close to 1. 
\item \textbf{Softmax}: The softmax activation function was used in the final layer to train the neural network under a binary cross-entropy loss\cite{christopher2016pattern}. 
\end{enumerate}

\subsection{Training Methodology}
\label{train}
\begin{figure}[htbp]
\includegraphics[width=0.48\textwidth]{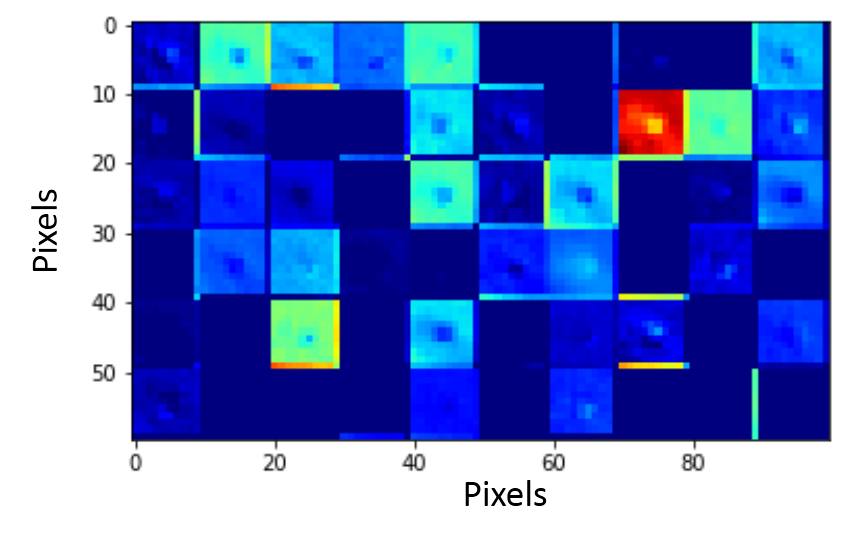}
\caption{Activation Map of the Convolution-1 layer depicting neuron activation upon detection of the plasmodium.}
\label{fig:activation}
\centering
\end{figure}

For the malaria and tuberculosis datasets, the network was trained using randomly initialized weights. As described in Section \ref{dset}, the intestinal parasite dataset has a very limited number of positive samples (162) which makes it difficult to train the network without causing over-fitting. Thus transfer learning was used. For layers in our network before Batch Normalization, weights from a Squeeze-Net pre-trained on the ImageNet database \cite{krizhevsky2012imagenet} were used. The model is trained using the Adam optimizer with a learning rate of \(1e^{-4}\) and batch size of 256\cite{kingma2014adam}. The same network architecture is used in all the cases. 
Training and initial performance estimation for the CNN model was performed using Keras\cite{chollet2015keras} framework on an Intel i5/6300 CPU. Fig. \ref{fig:activation} shows the activation map of the Convolution-1 layer in the Network for plasmodium detection on the malaria dataset. 

\subsection{Pathogen Detection}
Upon completion of CNN training, the resulting model was able to classify a sample as containing a pathogen of interest or not. In order to identify pathogens within an acquired image sample, we use the sliding window approach. In this approach, a sliding window of a fixed size traverses the image and queries the trained model to detect presence of the pathogen.
However, this approach produces multiple windows with high scores close to the correct location of object. Non-maximum Suppression is used with the aim of having one window per object within the test image\cite{rosten2006machine}. The procedure starts by selecting the best scoring window with the assumption that it covers an object of interest. Then, the windows that overlap with the selected window beyond a threshold are suppressed. The hyper-parameters: (1) overlap threshold (2) detection threshold are chosen based on design-space exploration to be 0.3 and 0.99 for our implementation respectively. 

\begin{figure}[htbp]
\centering
\includegraphics[scale=0.4]{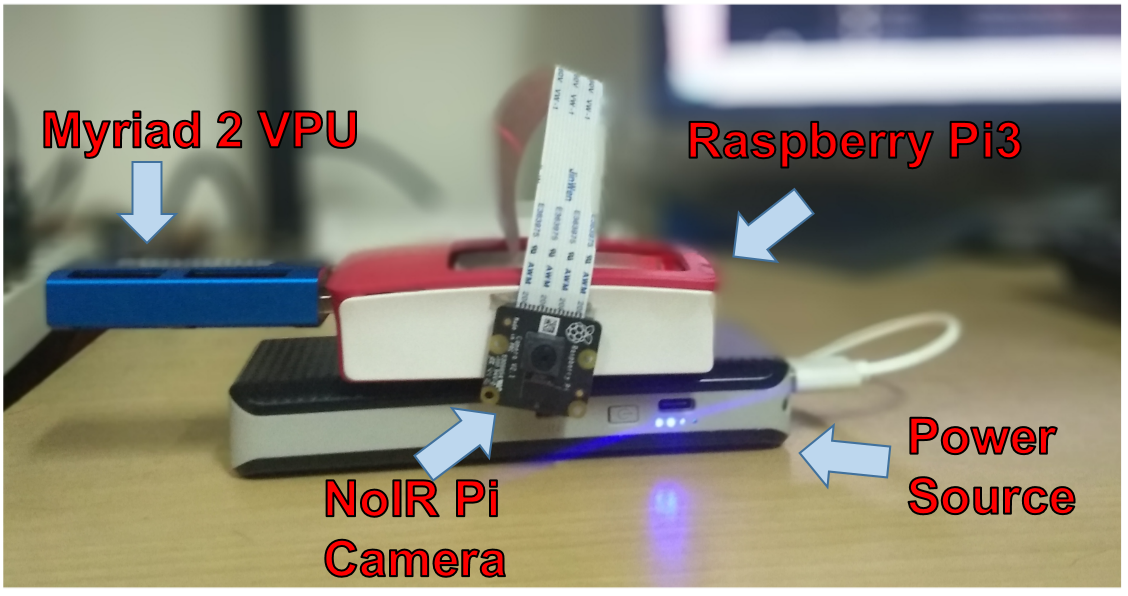}
\caption{Proposed Hardware Implementation}
\label{prop1}
\centering
\end{figure}
\subsection{Proposed dedicated Hardware Implementation}
The complete setup of the proposed POC solution is shown in Fig. \ref{prop1}. It includes Raspberry Pi 3, along with Myriad VPU (based on Movidius Neural Compute Stick). The Myriad VPU supports inference for networks designed and trained on common deep-learning frameworks such as TensorFlow\cite{abadi2016tensorflow} and Caffe\cite{jia2014caffe}. Caffe was used for training the proposed CNN model. VPU was attached to the Raspberry Pi over USB for offloading the trained CNN model to perform inference.

\section{Results and Discussion}
\label{res}
\subsection{Learning performance}

\begin{figure}[t]
\centering
\includegraphics[width=0.46\textwidth]{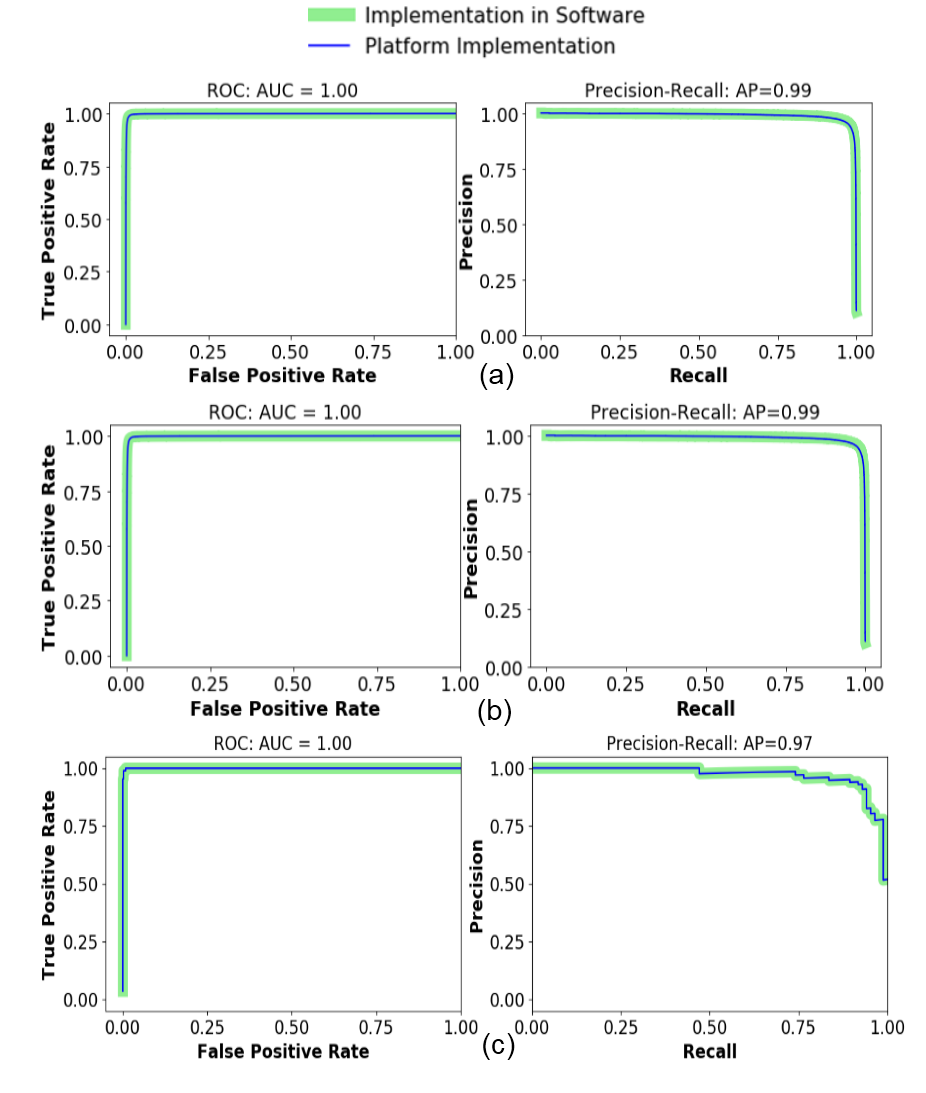}
\caption{ROC and precision-recall showing Area Under Curve (AUC) and Average Precision (AP) for implementation on Software(CPU) and Platform in each case (a) Malaria (b) Tuberculosis (c) Intestinal Parasites.}
\label{fig:roc}
\end{figure}

\begin{figure}[t]
\centering
\includegraphics[scale=0.3]{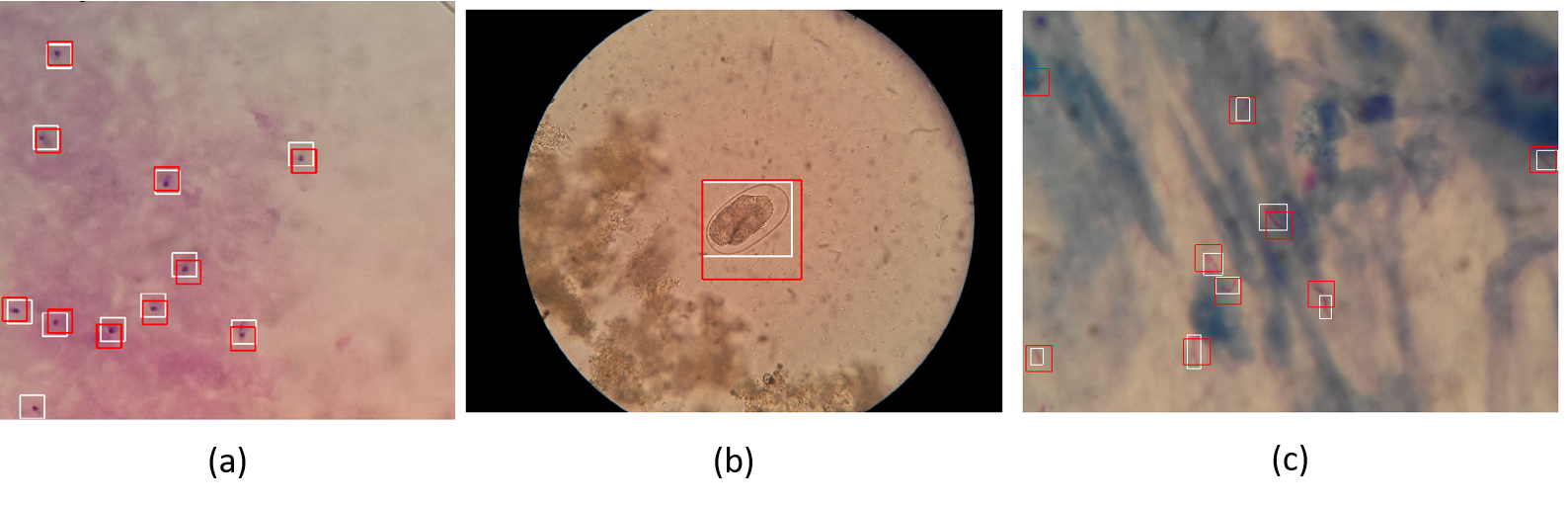}
\caption{Detected objects in test images (a) for plasmodium (b) for hookworm egg 
(c) for tuberculosis bacilli. White boxes show expert annotations on pathogen locations. Red boxes show detection by the proposed system.}
\label{fig:detection}
\centering
\end{figure}

The trained models are evaluated with corresponding test sets: for plasmodium detection with 290,555 test samples (9.8\% positive), for tuberculosis with 80262 samples (43.9\% positive), and for hookworm with 8585 samples (4.1\% positive). Receiver Operating Characteristics\cite{hajian2013receiver} and Precision-Recall curves are shown for both software (CPU) based and dedicated hardware (VPU) based CNN implementation in Fig \ref{fig:roc}. Best results achieved in software were replicated when implemented on the hardware platform. Fig \ref{fig:roc} shows overlapping curves for either case. Our proposed CNN was able to achieve an area greater than 0.99 under the receiver operating characteristic curve for all 3 datasets. 
Fig. \ref{fig:detection} shows that the annotations given in the dataset are close to those found by our proposed network.


\subsection{Memory Optimization}

\begin{figure}[htbp]
\includegraphics[scale=0.4]{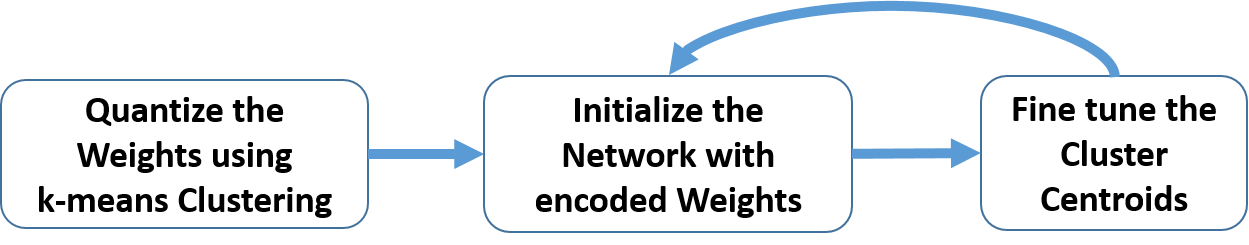}
\centering
\caption{Trained Quantization Method.}
\label{fig:trained}
\end{figure}

\begin{figure}[htbp]
\centering
\includegraphics[width=0.48\textwidth]{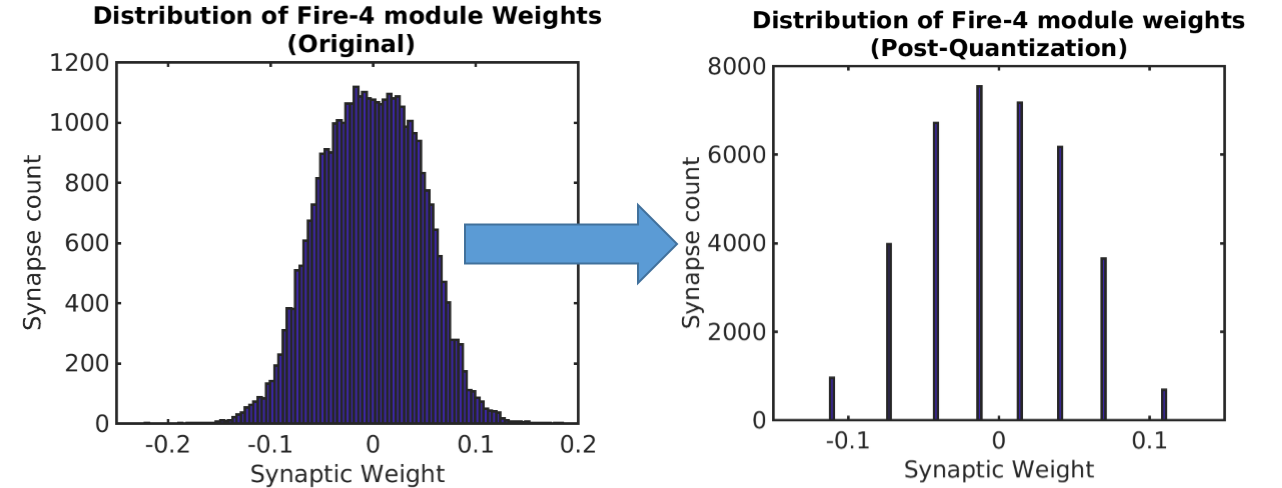}
\caption{Weight Quantization using K-Means Clustering on the Malaria Dataset.}
\label{fig:Quant}
\end{figure}

We used trained quantization and weight sharing to reduce the memory footprint of our proposed CNN model\cite{han2015deep}. 
In this method, the learned weights were quantized by applying the k-means clustering algorithm to each of the layers of the network separately as shown is Fig. \ref{fig:Quant}. The model is initialized with quantized weights, then fine-tuned again over the cluster centroids to regain accuracy as shown in Fig. \ref{fig:trained}. Table \ref{tab:memory} shows that the memory required by our implementation is 5x-30x less compared to previous implementations in literature \cite{quinn2016deep}.

\begin{table}[ht]
  \centering
  \caption{Comparison of Memory Requirements}
    \begin{tabular}{|P{1.4cm}|P{0.9cm}|P{1.7cm}|P{1cm}|P{1.4cm}|}
    \hline
    \multirow{3}{*}{\textbf{Dataset}} & \multicolumn{3}{c|}{\textbf{Weight Memory Size (kB)}} &  \\ \cline{2-4}
    & \textbf{Custom} & \textbf{Compression} & \textbf{Quinn} &\textbf{Compression}\\
    & \textbf{CNN} & \textbf{+ Custom CNN} & \textbf{et al.\cite{quinn2016deep}} & \textbf{Factor}\\
    \hline
    
    \textbf{Malaria} & 549.4 & 53.3 & 1548  & \textbf{29x} \\
    \hline
    \textbf{Tuberculosis} & 545.2 & 53.3 & 314.7 & \textbf{5.9x} \\
    \hline
    \textbf{Intestinal Parasites} & 629.2 & 133.2 & 1190  & \textbf{8.9x} \\
    \hline
    \end{tabular}%
  \label{tab:memory}%
\end{table}%

\subsection{Latency and Energy}
The average latency for the proposed CNN to perform 1000 inference operations for each dataset using the proposed hardware implementation was $\sim$ 2.7 s. Minimum latency was observed while using 6 SHAVE Vector Processors \cite{moloney20111tops,xu2017convolutional}. In order to compare the latency of inference operations on different platforms, we also tested the network running on a CPU: Intel Core i5-6300U @ 2.30GHz, 12 GB RAM. The power consumption of the proposed network on the hardware platform was measured during the inference stage by using a Keweisi KWS-V20 USB power monitor. For the proposed hardware implementation, the average power consumption for all inference networks was found to be $\sim$2.95 W which is significantly lower than the i5 CPU dissipation for the same task (shown in Table \ref{tab:analysis}). 
Power dissipation for the software implementation was estimated using the tool-powerstat \cite{powerstat}. 
\begin{table}[ht]
  \centering
  \caption{Network Performance on different hardware platforms}
    \begin{tabular}{|P{3.4cm}|P{2.2cm}|P{1.2cm}|}
    \hline
    \textbf{Platform } & \textbf{Movidius NCS + Raspberry Pi3} & \textbf{Intel i5/6300U} \\
    \hline
    \textbf{Time(/sample) (ms)} & 2.7  & 1.1 \\
    \hline
    \textbf{Power (W)} & 2.9 & 18.5 \\
    \hline
    \textbf{Energy Consumption (mJ)} & 8.0 & 20.4 \\
    \hline
  \end{tabular}%
  \label{tab:analysis}%
\end{table}

\section{Conclusion}
\label{conc}
In this paper, we demonstrate a low-power portable dedicated hardware-based solution for microscopy point of care diagnostic support. For demonstrating performance on real-world applications we used datasets for three different diseases. 
Based on design exploration results, the maximum memory requirement for inference network was found to be $\sim$ 133 kB.
The proposed solution had an inference latency of $\sim$ 3 ms/sample with a power consumption of $\sim$ 2.9 W which was 6x efficient compared to conventional CPU based implementation of the same CNN.
The memory and algorithmic optimization strategies used in the proposed design approach enable potential future deployment in rural areas and improve the access to diagnostic health-care in resource-constrained scenarios. Future work will involve integration with a digital microscope for the development of a low-cost packaged system for deployment.
\section*{Acknowledgements}
This research activity under the PI Prof. M. Suri is partially supported by the Department of Science \& Technology (DST), SERB-EMR, MHRD-Imprint (RP03417G), Government of India and IIT-D FIRP grants. 


\bibliographystyle{IEEEtran}
\bibliography{ref}



\end{document}